\pdfoutput=1

\documentclass[11pt]{article}

\usepackage[preprint]{coling}

\usepackage{times}
\usepackage{latexsym}

\usepackage[T1]{fontenc}

\usepackage[utf8]{inputenc}

\usepackage{microtype}

\usepackage{inconsolata}

\usepackage{graphicx}
\usepackage{rotating}
\usepackage{adjustbox}
\usepackage{amsmath}
\usepackage{multirow}
\usepackage{arydshln}
\usepackage{algorithm}
\usepackage{algpseudocode}

\title{Measuring Sample Importance in Data Pruning for Language Models based on Information Entropy}

\author{Minsang Kim \\
  Korea University \\
  Dept. of Computer Science and Engr. \\
  South Korea \\
  \texttt{kmswin1@korea.ac.kr} \\ \And
  Seungjun Baek \\
  Korea University \\
  Dept. of Computer Science and Engr. \\
  South Korea \\
  \texttt{sjbaek@korea.ac.kr} \\}

\begin{document}
\maketitle
\begin{abstract}
Compute-efficient training of language models has become an important issue. We consider data pruning for data-efficient training of LLMs. In this work, we consider a data pruning method based on information entropy. We propose that the samples in the training corpus be ranked in terms of their informativeness which we estimate through entropy functions.
The key idea is that, less informative samples are likely to contain redundant information, and thus should be pruned first. We use the entropy functions based on the negative log-likelihood and the average inverse word frequency of a sample as a surrogate to measure its informativeness. Experiments reveal that the proposed information-based pruning can improve upon various language modeling and downstream tasks, and enhance the generalization capability of language models. 
\end{abstract}

\section{Introduction \& Related work}
Pretraining large language models~(LLM) requires high computational costs due to their sizes and commensurately large datasets~\cite{kaplan2020scaling, hoffmann2022training}. Recently, computationally efficient methods to train deep models based on \emph{data pruning}  have gained interest. Data pruning concerns deciding which samples are important for training, and removing unimportant samples from training datasets. 
\cite{sorscher2022beyond} proposed that neural scaling law~\cite{kaplan2020scaling} may be overcome with data pruning. Their method uses distances between samples and their cluster centroids in the embedding space to measure the sample importance. \cite{tan2024data} assessed the sample importance by measuring the change in empirical risk when the sample is removed from the training set. 
 

The aforementioned works apply data pruning to labeled data for supervised learning or focus on visual understanding tasks. However, data pruning for LLMs has been relatively underexplored. Text deduplication methods have been proposed ~\cite{raffel2020exploring, abbas2023semdedup} to remove redundant data based on semantic similarity. However, they do not quantify the importance of samples for data pruning. Recently, \cite{marion2023less, ankner2024perplexed} proposed data pruning methods based on {perplexity}. They used perplexity to measure the difficulty of predicting next token, and observed that removing easy samples from training datasets tends to improve performances. While this method is partly related to ours, we take a different approach based on information entropy which achieves higher data efficiency.
Recently, connections between LLMs and data compression have been explored. \cite{ge2023context} proposed In-Context Autoencoder for context compression. \cite{deletang2023language} extensively studied the view of LLM as a model for data compression. The predictive power of LLMs can be used for an optimally efficient expression of data based on the prediction probability, to compress various types of multi-modal data.

In this paper, we take an information-entropic view on data pruning. The (information) entropy of text samples quantifies how many bits are required to efficiently describe, i.e., compress, the samples. We use entropy functions to measure how informative samples are, and prune uninformative samples from the dataset. We consider two metrics: firstly, the negative log-likelihood of samples which can be interpreted as the amount of information required to express the samples. Secondly, an entropy function based on the inverse frequency of words which quantifies the rarity or \emph{surprisal} of individual words irrespective of their contexts.
We propose to combine two metrics as a surrogate to measure the sample importance.

To implement the proposed pruning, we first train \emph{data probe model} which is a small model trained with a subset of the corpus. 
Next, we prune the dataset based on the negative log-likelihood estimates by the data probe model and the rarity of words based on mean inverse frequency. Finally, we train a target model with the pruned dataset. Experimental results demonstrate that, the proposed pruning can actually \emph{improve} the model performance compared to both random and perplexity-based pruning. In some cases, the performances of generation and/or downstream tasks are maintained up to pruning 50\% of the pretraining corpus. The key insight is that the proposed pruning based on sample information helps remove redundant or less helpful data, which enables compute-efficient training and enhances the generalization capability of language models.

\section{Method}
\subsection{Measures of Sample Importance}
We take an information theoretic view on data pruning for language models as follows. 
Let $p(\cdot)$ denote the true distribution of words from a corpus. Given a sample $W=(w_1,w_2,\ldots, w_n)$ which is a sequence of words $\{w_i\}$, consider
\begin{align}
H(W,p) = \frac{1}{n}\sum_{i=1}^n \log\frac{1}{p(w_i | w_{<i})}\label{eq:entropy}
\end{align}
$H(W,p)$ can be regarded as the minimum description length required to compress words in $W$. Indeed, the optimal compression based on entropy, e.g., arithmetic coding \cite{rissanen1976generalized}, uses approximately \(\lceil\log_21/p(w_i | w_{<i})\rceil\) bits to encode $w_i$ given previous words. Thus, $H(W,p)$ represents the amount of information per word in passage $W$.  $H(W,p)$ also represents the \emph{surprisal} of sample $W$ averaged over its words, where the surprisal refers to the amount of information or uncertainty in the information-theoretic context.

Consider a next-word predictor $q$, i.e., a language model (LM). The negative log-likelihood of $W$ on model $q$ is given by 
\begin{align}
H(W,q) 
 = \frac{1}{n}\sum_{i=1}^n \log\frac{1}{q(w_i | w_{<i})}\label{eq:ll}
\end{align} 
From an information-theoretic view, \eqref{eq:ll} is the mean description length of words in $W$ using sub-optimal distribution $q$. If we average $nH(W,q)$ over $W$, it is the \emph{cross-entropy} or \emph{log-loss} used for training the LM. Thus, training is equivalent to minimizing the description length of, i.e., compressing, the corpus data \cite{deletang2023language}. 
Thus, if $q$ is well-trained, \eqref{eq:ll} is close to \eqref{eq:entropy}, and the 
 information content of $W$ can be approximately measured by \eqref{eq:ll}. In that case, $H$ is likely to be large for samples with large surprisal, i.e., highly informative samples. By contrast, samples containing redundant or repetitive passages will be relatively easy to compress, and $H$ is likely to be small. In this work, \eqref{eq:ll} is used as one of two measures of sample importance for data pruning.

Next, we consider additional measures of sample importance based on the word frequency. Consider distribution $f(w_1,...,w_n) = \prod_{i=1}^n \hat f(w_i)$ where $\hat f(w_i)$ denotes the (normalized) frequency of the occurrence of $w_i$ in the corpus. Specifically, $f$ is the likelihood of sample $W$ assuming the words in $W$ occur independently irrespective of their order as in, e.g., the Bag-of-Words model. The number of bits per word required to express $W$ assuming such independence (albeit having suboptimal length) 
is
\begin{align}
H(W, f) = \frac{1}{n}\log\frac{1}{f(w_1,...,w_n)}=\frac{1}{n}\sum_{i=1}^n \log\frac{1}{\hat f(w_i)}\nonumber 
\end{align}
$H(W,f)$ quantifies how rare the words in $W$ are, i.e., $\log \hat f(w_i)^{-1}$ is the surprisal of $w_i$ based on its rarity. As previously, averaging $nH(W,f)$ over $W$ gives the cross-entropy between $p$ and $f$. Unlike \eqref{eq:ll} which measures the surprisal of words conditional on previous context, $H(W,f)$ independently evaluates each word and emphasizes the importance of samples with rare terms. Indeed, samples containing infrequent words, e.g., technical or medical terms, can be useful for expanding the LLM's knowledge.

Finally, the proposed heuristic for measuring sample importance of $W$ with model $q$ is 
\begin{align}
H(W,q) + H(W,f)    \label{eq:all}
\end{align}
This captures the information content of $W$ measured by autoregressive model $q$  with additional emphasis on the rarity of words in the corpus. The first term of \eqref{eq:all} quantifies the average surprisal in a context-dependent manner, while the second term does so in a context-free manner.

\paragraph{Another Interpretation.} One can show that the \emph{perplexity} of $W$, denoted by $\textrm{Perp}(W)$, is given by \(
e^{H(W,q)}.
\)  
Recent works~\cite{marion2023less, ankner2024perplexed} showed that data pruning based solely on \(\textrm{Perp}(W)\) can be effective. Those observations are consistent with our framework such that \eqref{eq:ll} can be used to sort out informative samples. Meanwhile, the proposed measure in \eqref{eq:all} implies that, the samples are ranked according to 
\[
e^{H(W,q) + H(W,\hat q)} = \left(\prod_{i=1}^n f(w_i)^{-1}\right)^{\frac{1}{n}}\cdot\textrm{Perp}(W)
\]
Thus, \eqref{eq:all} can be viewed as \emph{weighted perplexity} where the weight is the geometric mean of the inverse-frequency of words. In conclusion, sample importance \eqref{eq:all} captures the perplexity of autoregressive estimation weighted by the additional surprisal: the inverse word frequency.

\subsection{Pruning Method}\label{method}
\noindent\textbf{Overview}. We first train reference model $q$, called \emph{data probe model}. The goal is to economically measure $H(\cdot,q)$ of the samples. 
Also,  all the word/token frequencies are pre-computed for $H(\cdot, f)$. The samples with low $H(\cdot,q) + H(\cdot, f)$ are pruned from the dataset, and the pruned dataset is used for training the target LM.

\noindent\textbf{Data probe model.} 
To reduce computational costs, we train the data probe model $q$ on a subsample of the pre-training corpus. Still, the subsample size should be sufficient so that $H(\cdot,q)$ provides a rough estimate of the sample information which is used as a relative measure for comparing sample importance.
We heuristically determine the subsample size by hypothesizing that the model is sufficiently trained if the rate of decrease in the training loss saturates, similar to the early stopping method.

In our experiments, the subsample size is set to be about 12\% of the training dataset: see Fig. \ref{fig:train_probe} in appendix~\ref{appendix:training_probe}. 
Later experiments show that the proposed level of training is sufficient for a good pruning performance.

\noindent\textbf{Pruning.} We will denote the training dataset by $\mathcal D$ and the target pruning ratio by  $\eta$. We first rank all the samples $W\in \mathcal D$ by  $H(W,q) + H(W, f)$, and then prune bottom $\eta$\% from $\mathcal D$. The pruned dataset is used for training the target LM. 

\section{Experiment}
\subsection{Experimental Settings}
As a pretraining corpus, we use a subset of c4~\cite{raffel2020exploring} via random sampling, which is a total of about 3 billion tokens from GPT-2 tokenizer~\cite{radford2019language}. We use decoder-only transformers~\cite{vaswani2017attention} as the language model. As the data probe model, we use a 125M-parameter model trained on about 0.3 billion tokens randomly sampled from the total corpus. As the target model, we use both 125M- and 345M-parameter models. We evaluate the language modeling of target models on One billion words~\cite{chelba2014one} and wikitext-103~\cite{merity2016pointer} where the models are pre-trained on different datasets. The downstream tasks for the target models are evaluated using glue benchmark~\cite{wang2018glue}. Detailed hyperparameters are in Appendix~\ref{hyperparameters}.

\begin{table}[t!]
\begin{adjustbox}{width=.4\textwidth, center}


\begin{tabular}{|l|lll|}
\hline
\textbf{Pruning \%} & \multicolumn{1}{l|}{\textbf{Proposed}} & \multicolumn{1}{l|}{\textbf{Perplexity}} & \textbf{Random} \\ \hline
0                   & \multicolumn{3}{c|}{90.23}                                                                          \\ \hline
10                  & \multicolumn{1}{l|}{\underline{\textbf{83.48}}}    & \multicolumn{1}{l|}{\underline{89.32}}               & 90.95           \\
20                  & \multicolumn{1}{l|}{\underline{\textbf{83.76}}}    & \multicolumn{1}{l|}{\underline{86.88}}               & 93.40           \\
30                  & \multicolumn{1}{l|}{\underline{\textbf{84.07}}}    & \multicolumn{1}{l|}{\underline{87.09}}               & 99.89           \\
40                  & \multicolumn{1}{l|}{\underline{\textbf{84.03}}}    & \multicolumn{1}{l|}{\underline{84.19}}               & 93.73           \\
50                  & \multicolumn{1}{l|}{\underline{\textbf{84.13}}}    & \multicolumn{1}{l|}{91.11}               & 97.08           \\
60                  & \multicolumn{1}{l|}{\underline{\textbf{87.33}}}    & \multicolumn{1}{l|}{91.24}               & 101.32          \\
70                  & \multicolumn{1}{l|}{\textbf{92.32}}    & \multicolumn{1}{l|}{102.93}              & 104.65          \\
80                  & \multicolumn{1}{l|}{\textbf{100.37}}   & \multicolumn{1}{l|}{101.70}              & 110.39          \\
90                  & \multicolumn{1}{l|}{\textbf{138.72}}   & \multicolumn{1}{l|}{147.18}              & 165.39          \\ \hline
\end{tabular}

\end{adjustbox}
\caption{Perplexity of GPT-125M per pruning ratio on One Billion Words corpus. \textbf{Bold} denotes the best performance for each pruning ratio, and the \underline{underline} indicates better performances than no-pruning case.}
\label{tab:lm1b}
\end{table}

\begin{table}[t!]
\begin{adjustbox}{width=.4\textwidth, center}

\begin{tabular}{|l|lll|}
\hline
\textbf{Pruning \%} & \multicolumn{1}{l|}{\textbf{Proposed}}                       & \multicolumn{1}{l|}{\textbf{Perplexity}}            & \textbf{Random}                \\ \hline
0                   & \multicolumn{3}{c|}{52.23}                                                                                                                          \\ \hline
10                  & \multicolumn{1}{l|}{\textbf{\underline{50.19}}} & \multicolumn{1}{l|}{\underline{50.59}} & \underline{51.56} \\
20                  & \multicolumn{1}{l|}{\textbf{\underline{48.43}}} & \multicolumn{1}{l|}{\underline{51.71}} & 52.25                          \\
30                  & \multicolumn{1}{l|}{\underline{\textbf{49.81}}}                          & \multicolumn{1}{l|}{53.46}                          & 54.97                          \\
40                  & \multicolumn{1}{l|}{\underline{\textbf{51.44}}}                          & \multicolumn{1}{l|}{55.06}                          & 56.91                          \\
50                  & \multicolumn{1}{l|}{\textbf{54.22}}                          & \multicolumn{1}{l|}{58.26}                          & 59.63                          \\
60                  & \multicolumn{1}{l|}{\textbf{55.69}}                          & \multicolumn{1}{l|}{61.97}                          & 64.29                          \\
70                  & \multicolumn{1}{l|}{\textbf{55.80}}                          & \multicolumn{1}{l|}{62.66}                          & 70.22                          \\
80                  & \multicolumn{1}{l|}{\textbf{64.41}}                          & \multicolumn{1}{l|}{71.46}                          & 85.42                          \\
90                  & \multicolumn{1}{l|}{\textbf{90.21}}                          & \multicolumn{1}{l|}{115.03}                         & 135.34                         \\ \hline
\end{tabular}

\end{adjustbox}
\caption{Perplexity of GPT-125M per pruning ratio on wikitext-103 corpus. \textbf{Bold} denotes the best performance for each pruning ratio, and the \underline{underline} indicates better performances than no-pruning case.}
\label{tab:wikitext103}
\end{table}

\begin{figure*}[t!]
    \centering
    \includegraphics[width=1.0\textwidth]{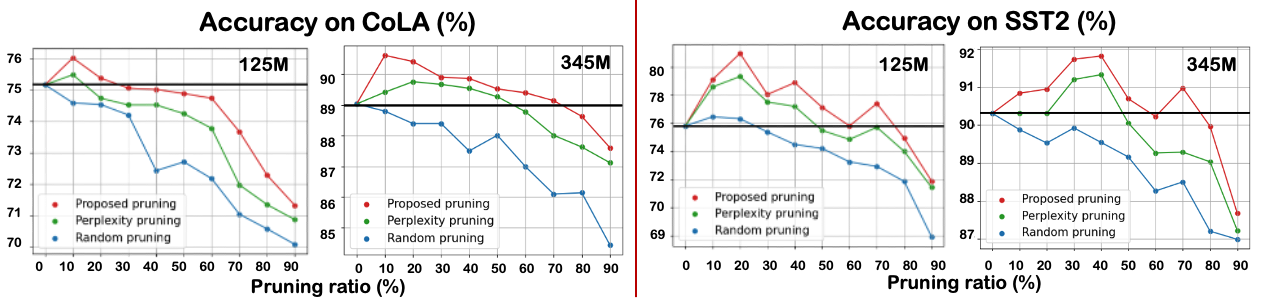}
    \caption{Left: Text classification results on CoLA. Right: Text similarity results on SST2. The black line indicates the performance without pruning.}
    \label{fig:downstream}
\end{figure*}

\subsection{Language Modeling}
We evaluate language modeling of target models using two test corpus. 
Table~\ref{tab:lm1b} shows the perplexity of each model per pruning ratio on One Billion Words corpus \cite{chelba2014one}. The proposed pruning significantly outperforms the random pruning. 
Surprisingly, the proposed pruning achieves \emph{better} performance up to 60\% pruning ratio as compared to the no-pruning case and outperforms perplexity-based and random data pruning in all pruning ratios by a large margin. We observe a similar trend with different corpus: Table~\ref{tab:wikitext103} shows the perplexity results of language modeling on wikitext-103 corpus~\cite{merity2016pointer}.

The results show that, our pruning can improve the generalization capability of the target model, which we provide the following as a possible explanation. Our pruning tends to remove samples that are less informative or contain highly redundant information. Such redundancy in the dataset may cause overfitting such samples similar to duplication, i.e., the model may get biased towards such information, hampering generalization. Thus, pruning based on estimates of the information content of samples may alleviate overfitting problems.

In addition, the proposed approach of combining informativeness and inverse word frequency to evaluate the sample importance seems effective. For similarly informative samples, i.e., samples with similar $H(\cdot,q)$, our pruning prioritizes samples containing infrequent terms, i.e., larger $H(\cdot, f)$. Table~\ref{tab:lm1b} shows comparisons with the pruning method based only on perplexity, e.g., \cite{marion2023less, ankner2024perplexed}. The results show that our method has better generalization capabilities than perplexity-only pruning.

\subsection{Downstream tasks}
We finetune and evaluate each model in downstream tasks such as text classification and textual similarity tasks from glue~\cite{wang2018glue}. The left of Fig.~\ref{fig:downstream} shows the text classification results on CoLA dataset, which compares  the proposed method, perplexity pruning, and random pruning. The results show that, the accuracy of 125M trained with our data pruning is always above that of the perplexity and random pruning. 
Next, we scaled up the target model to 345M. Fig.~\ref{fig:downstream} shows that the proposed data pruning method achieves significant performance gains over the other methods. Notably, the proposed pruning outperforms {no-pruning} case with almost up to 50\% pruning ratio, i.e., pruning helps improve the performance of language models.

The right of Fig.~\ref{fig:downstream} illustrates the results of the textual similarity task on SST2 dataset for 125M and 345M models. For both models, while the performance continually drops with random pruning, the proposed method achieves similar or better accuracy than no-pruning case up to 50\% pruning ratio. In particular, the proposed method obtains higher accuracy in all pruning ratios compared to the other pruning methods.

Overall, we make a similar observation to the pretraining case: the proposed pruning not only reduces the training cost of language models, but also \emph{improves} the performance of downstream tasks. We conclude that pruning based on the combination of informativeness and word-frequency of samples can improve the generalization capabilities of the models on downstream tasks.

\section{Conclusion}
In this paper, we took an information-theoretic view of data pruning for training language models. The negative log-likelihood and the mean inverse word frequency were combined to measure the sample importance. 
Pruning based on the proposed 
importance enabled removing samples with redundant information, which not only reduces computational costs but also, surprisingly, enhances the generalization capability of language models. Such effectiveness of our pruning was observed and validated at experiments with various corpus and downstream tasks.

\section{Limitations}
Due to the nature of our work, we needed to train language models from scratch. Given the limited resources, our experiments were conducted for the models of size 125M and 345M parameters, and not for billion-scale models.
Nevertheless, we showed that pruning by a smaller probe model (GPT-125M) improves the generalization performance of larger models (GPT-345M). 
In addition, the initial model and dataset sizes in our pruning experiments 
is set to be Chinchilla-optimal \cite{hoffmann2022training}. Assuming that such neural scaling laws for model and dataset sizes are used, the proposed pruning is expected to remain effective with up-scaling, i.e., for larger language models.

\bibliography{custom}

\appendix

\clearpage
\onecolumn
\section{Appendix}

\subsection{Loss curve of Data Probe Model}\label{appendix:training_probe}
\begin{figure}[h!]
    \centering
    \includegraphics[width=.45\textwidth]{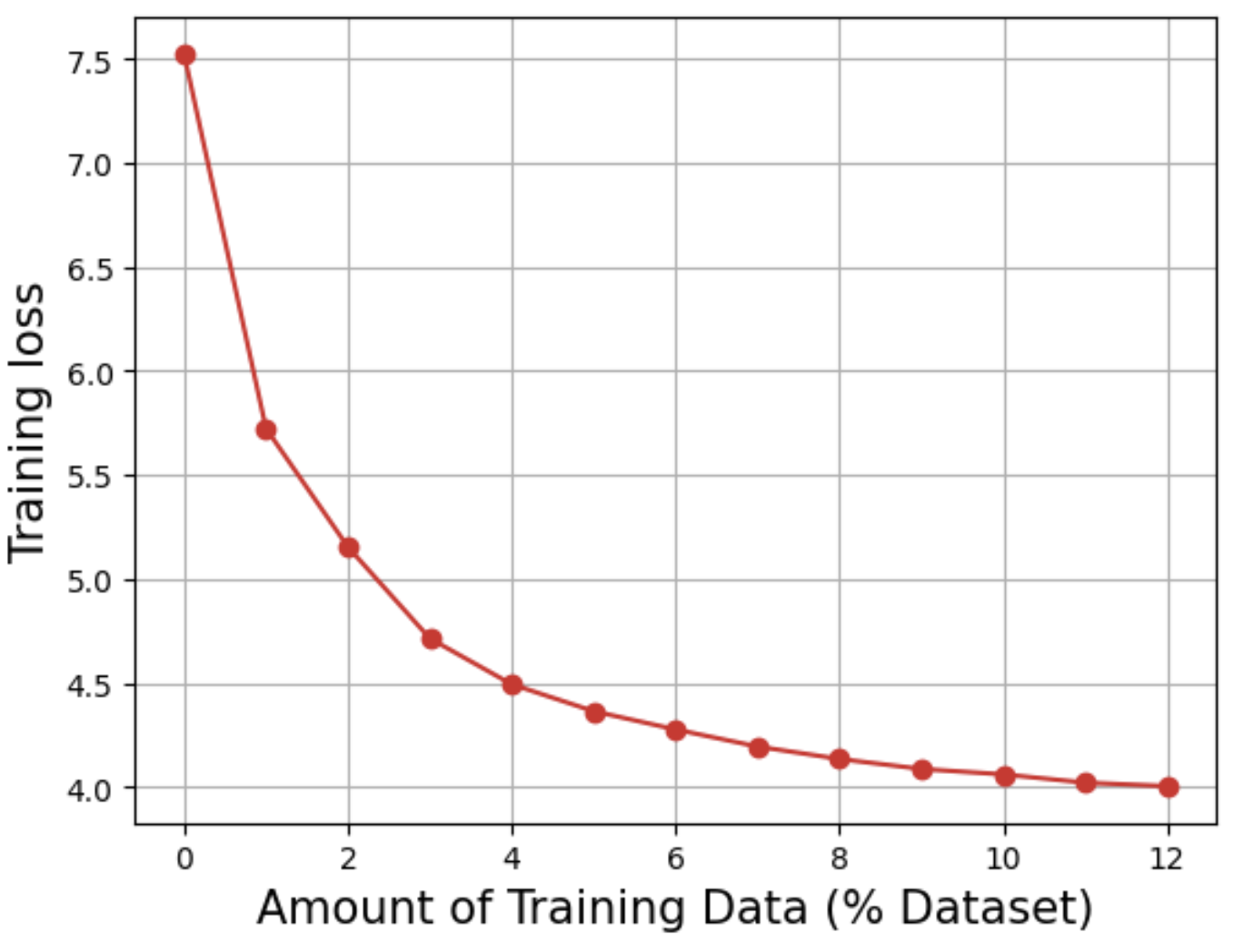}
    \caption{Loss curve of GPT-125M over 1 sweep of the training dataset. For data efficiency, we stopped training the probe model at the point where the decrease in loss saturates, i.e., at about 12\% of the entire dataset.
    }
    \label{fig:train_probe}
\end{figure}

Fig.~\ref{fig:train_probe} shows the loss curve of the data probe model. We stopped training at about 12\% of the dataset for the probe model since training the entire corpus is not efficient in practice. We heuristically determine the subsample size by hypothesizing that the model is sufficiently trained if the rate of decrease in the training loss saturates, similar to the early stopping method.

\subsection{Pretraining Loss}\label{appendix:pretrain_loss}

\begin{figure}[h!]
    \centering
    \includegraphics[width=.8\textwidth]{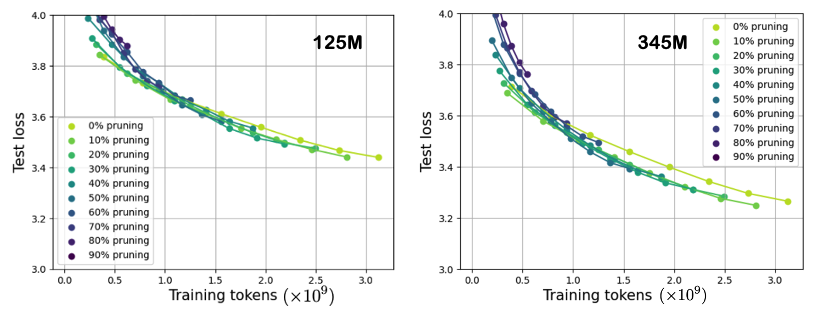}
    \caption{Left: Test loss of 125M GPT per training token. Right: Test loss of 345M GPT per training token.}
    \label{fig:loss}
\end{figure}

We first experimented with various pruning ratios of low entropy samples of the same size (125M) as the target model. The left of Fig.~\ref{fig:loss} illustrates that applying data pruning models converges faster in test loss. Even, pruning 10\% achieves lower test loss than training the entire data. This shows that low-entropy data is less helpful for training LLMs and pruning low-entropy samples can train LLMs compute-efficiently.

In addition, the Right of Fig~\ref{fig:loss} shows the test loss of the 345M target model during pretraining. Interestingly, pruning 10\% achieves a \emph{lower} and pruning 20\% obtains a similar test loss, as compared to the no-pruning case. As we will see later, the lower loss indeed leads to improved performances in language modeling and downstream tasks with pruning.

\clearpage

\subsection{Qualitative Analysis}
\begin{table}[h!]
\centering
    \begin{adjustbox}{width=1.1\textwidth,center}
    \begin{tabular}{|l|}
    \hline
    \textbf{Sample cases} \\ \hline
    \multicolumn{1}{|p{\textwidth}|}{
\small
... \newline
What kind of experts are wondering about the new spacesuit? \newline
Who designed the new \textbf{spacesuit}?
What will the spacesuit help \textbf{astronauts} deal with the call of? \newline
Where won't the new \textbf{spacecraft} be able to go to?
...
}                \\ \hline
    \multicolumn{1}{|p{\textwidth}|}{
\small
... \newline
\textbf{GeneSigDB} Published Gene Signatures PubMedIDs of publications reporting gene signatures containing \textbf{SAMD12} from the GeneSigDB Published Gene Signatures dataset.
GEO Signatures of Differentially Expressed Genes for Gene Perturbations gene perturbations changing expression of \textbf{SAMD12} gene from the GEO Signatures of Differentially Expressed Genes for Gene Perturbations dataset. \newline
...
}            \\ \hline
    \multicolumn{1}{|p{\textwidth}|}{
\small
... \newline
\textbf{ACE Inhibitors}: Enhanced hypotensive effect when given with diuretics. A marked fall in blood pressure and deterioration in renal function may be seen when ACE inhibitors are added to furosemide therapy. The dose of furosemide should be reduced for at least three days, or the drug stopped, before initiating the ACE inhibitor or increasing the dose of an ACE inhibitor. \newline
\textbf{Alpha-blockers}: Enhanced hypotensive effect when diuretics are given with alpha-blockers, also increased risk of first dose hypotension with post-synaptic alpha-blockers such as prazosin. \newline
\textbf{Analgesics}: Diuretics can increase the risk of \textbf{nephrotoxicity} of NSAIDs, also antagonism of diuretic effect. \textbf{Antagonism} of diuretic effect (especially with \textbf{indomethacin} and \textbf{ketorolac}). \textbf{Salicylic} toxicity may be increased by furosemide. \newline
\textbf{Angiotensin –II Receptor Antagonists}: Enhanced hypotensive effect when diuretics given with angiotensin-II receptor antagonists. \newline
...
}            \\ \hline
\end{tabular}
\end{adjustbox}
\caption{Samples of the mean word rarity is much larger than negative log-likelihood.}
\label{tab:case}
\end{table}

Table~\ref{tab:case} shows the mean inverse word frequency is much larger than the negative log-likelihood of the data probe model. The cases show that the proposed method can measure high-importance scores by space, genetic, or medical terminologies. These samples show that our proposed method assigns a high importance score to rare word samples, even though language models measure a low perplexity.

\subsection{Hyperparameters}\label{hyperparameters}
\begin{table}[h!]
    \centering
    \begin{adjustbox}{width=0.7\textwidth,center}
    \begin{tabular}{|c|c|}
        \hline
        \textbf{Hyperparameter} & \textbf{Value}  \\
        \hline
        Model size & 125M, 345M \\
        Pretrain learning rate & 5e-4 \\
        Pretrain Learning rate scheduler & Cosine \\
        Pretrain warmup ratio & 1\% \\
        Pretrain weight decay & 0.1 \\
        Pretrain batch size & 512 \\
        Sequence length & 1024 \\
        Finetuning learning rate & 2e-5 \\
        Finetuning weight decay & 0.01 \\
        Finetuning batch size & 32 \\
        Optimizer & Adam~\cite{kingma2014adam} \\
       \hline
    \end{tabular}
    \end{adjustbox}
    \caption{Detailed hyperparameters.} 
    \label{tab:hyper}
\end{table}

\end{document}